\documentclass[conference]{IEEEtran}
\IEEEoverridecommandlockouts
\usepackage{cite}
\usepackage{amsmath,amssymb,amsfonts}
\usepackage{algorithmic}
\usepackage{graphicx}
\usepackage{textcomp}
\usepackage{xcolor}
\usepackage{comment}
\usepackage{subfigure}
\def\BibTeX{{\rm B\kern-.05em{\sc i\kern-.025em b}\kern-.08em
    T\kern-.1667em\lower.7ex\hbox{E}\kern-.125emX}}
\begin{document}


\def \ie{\textit{i.e.},}
\def \eg{\textit{e.g.},}
\def \etal{\textit{et al.},}

\title{Delving into the Scale Variance Problem in Object Detection}


\author{
\IEEEauthorblockN{Junliang Chen, Xiaodong Zhao and Linlin Shen*\thanks{*: Corresponding author}}
\IEEEauthorblockA{Computer Vision Institute, School of Computer Science and Software Engineering, Shenzhen University, China, \and Shenzhen Institute of Artificial Intelligence of Robotics of Society, Shenzhen, China, \and Guangdong Key Laboratory of Intelligent Information Processing, Shenzhen University, Shenzhen 518060, China\\
\{chenjunliang2016, zhaoxiaodong2020\}@email.szu.edu.cn, llshen@szu.edu.cn}
}



\maketitle

\begin{abstract}
   Object detection has made substantial progress in the last decade, due to the capability of convolution in extracting local context of objects. However, the scales of objects are diverse and current convolution can only process single-scale input. The capability of traditional convolution with a fixed receptive field in dealing with such a scale variance problem, is thus limited. Multi-scale feature representation has been proven to be an effective way to mitigate the scale variance problem. Recent researches mainly adopt partial connection with certain scales, or aggregate features from all scales and focus on the global information across the scales. However, the information across spatial and depth dimensions is ignored. Inspired by this, we propose the multi-scale convolution (MSConv) to handle this problem. Taking into consideration scale, spatial and depth information at the same time, MSConv is able to process multi-scale input more comprehensively. MSConv is effective and computationally efficient, with only a small increase of computational cost. For most of the single-stage object detectors, replacing the traditional convolutions with MSConvs in the detection head can bring more than 2.5\% improvement in AP (on COCO 2017 dataset), with only 3\% increase of FLOPs. MSConv is also flexible and effective for two-stage object detectors. When extended to the mainstream two-stage object detectors, MSConv can bring up to 3.0\% improvement in AP. Our best model under single-scale testing achieves 48.9\% AP on COCO 2017 \textit{test-dev} split, which surpasses many state-of-the-art methods.
\end{abstract}

\begin{IEEEkeywords}
object detection, scale variance, multi-scale convolution
\end{IEEEkeywords}

\begin{figure}[h]
    \centering
    \includegraphics[width=0.46\textwidth]{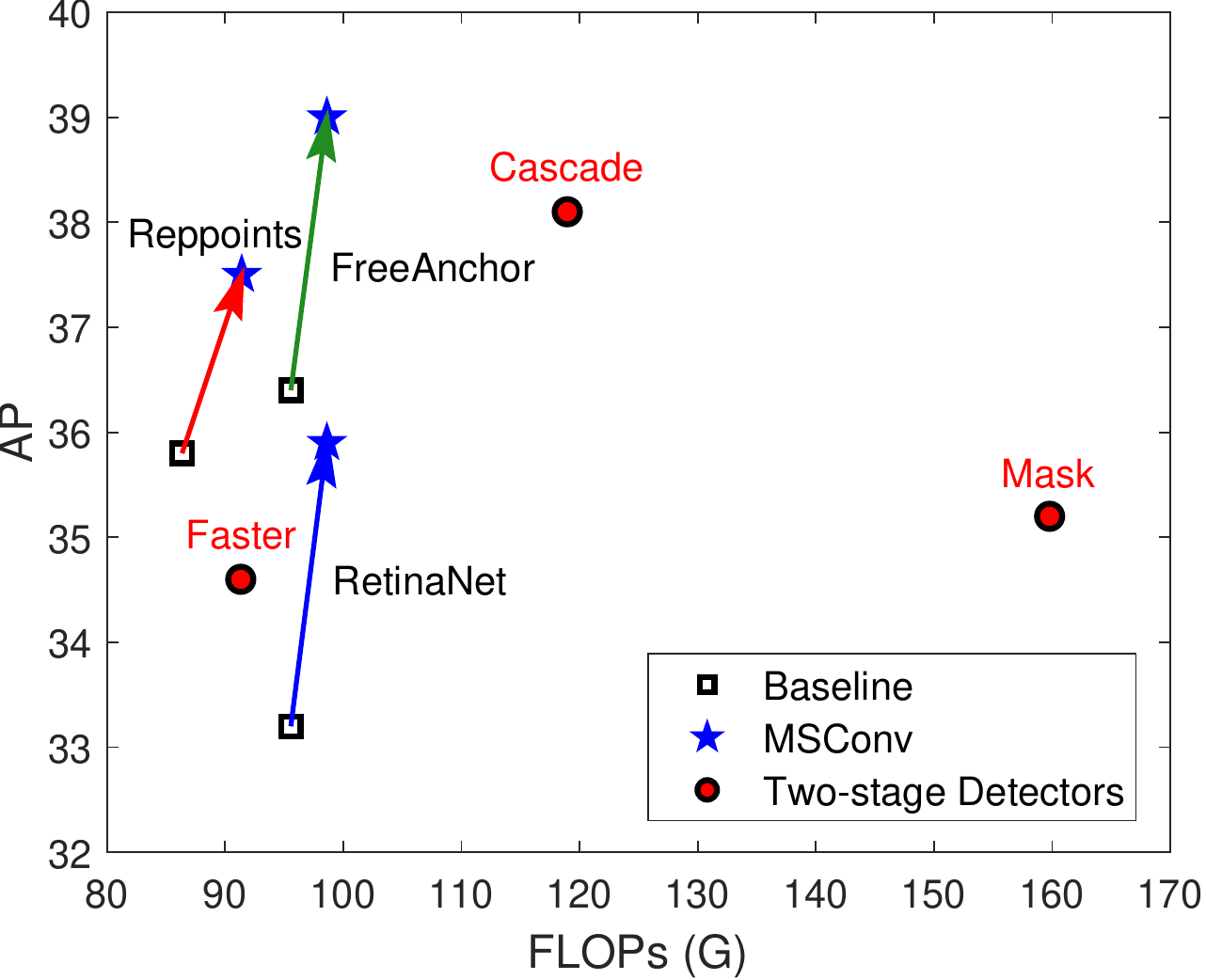}
    \caption{Performance on COCO \textit{val-2017} split of multi-scale convolution in various single-stage detectors including anchor-based FreeAnchor \cite{zhang2019freeanchor} and anchor-free RepPoints \cite{yang2019reppoints}. Two-stage detectors like Faster R-CNN (Faster) \cite{renNIPS15fasterrcnn}, Mask R-CNN (Mask) \cite{He_2017_ICCV}, and Cascade R-CNN (Cascade) \cite{cai18cascadercnn} are provided for reference. Our MSConv can significantly improve the APs of different detectors. All the models are trained on ResNet-50 \cite{He_2016_CVPR} backbone with a resolution of $640 \times 640$.}
    \label{fig:performance}
\end{figure}

\section{Introduction}

Object detection is a fundamental challenge in computer vision, which contains two sub tasks: location and classification. For any input image, the object detector is supposed to find out the position and category of all the objects available in the image. Unlike object recognition which only requires classification information, in object detection, we need to obtain the features containing accurate scale information of objects to locate them. The scale of different objects may vary in a wide range, making it difficult to represent large and small objects using the same scale. To alleviate the scale variance problem, researchers have made a lot of attempt.

In the earlier researches, image pyramid is an effective idea to solve the scale variances. The input image is resized to different resolutions, resulting in an image pyramid. Hand-engineered features are then extracted on the image pyramid. In image recognition, with the development of convolutional neuron networks (CNN), hand-engineered features are gradually replaced by the features computed by CNN. CNN is more robust to the scale variance and translation, thus improves the recognition performance based on a single image. Many recent top methods in the ImageNet \cite{5206848} and COCO \cite{10.1007/978-3-319-10602-1_48} detection challenges have utilized multi-scale testing to extract features from an image pyramid using CNN. Each level of the image pyramid is passed through the same CNN, to generate features at different scales. The idea to use CNN to extract features from an image pyramid gives a multi-scale feature representation of the original image. As all the levels of the image pyramid are passed through the whole network, the relative features are semantically strong, including the finest features with the highest resolution.

Image pyramid is a simple and good solution to represent the image at different scales. However, it is time-consuming, due to repeatedly network forward on different level of the image pyramid. To alleviate this problem, researchers have tried to make full use of the inherent characteristic of deep CNNs. Modern deep CNNs usually consist of many layers, including down-sample layers which generate features with decreased resolutions, such as poolings and strided convolutions. Given an input image, the deep CNN generates features with different scales, \ie a feature hierarchy. As the depth goes deeper, the corresponding layer becomes semantically stronger. Therefore, the last layer of the network is representative and widely used for object detection (\eg YOLOv1 \cite{yolov1}).

However, the semantically strongest features from the CNN have the lowest resolution, which reduces the spatial representation capability. Besides, the intrinsic feature hierarchy brings large semantic gap between the highest and lowest level. Thus, to some extent, it is limited to only use the features with strongest semantic for object classification and location. To solve this issue, the Single Shot MultiBox Detector (SSD) \cite{liu2016ssd} first attempts to utilize the existing feature hierarchy generated by the CNN. SSD takes the pyramidal features as inputs, and conducts prediction independently on each level of the pyramid. In consideration of realtime processing, SSD builds up detection heads from the high-level layers (\eg conv4\_3 layer of VGG-16 network \cite{Simonyan_2015_ICLR}) with low resolutions, which can not represent small objects well.

To make better use of the diverse semantics of features from different scales. Feature Pyramid Networks (FPN) explores the connection patterns between the multiple layers. FPN proposes the lateral connection of two adjacent layers in a top-down manner, and takes advantage of the representation capability of the high-resolution layers (\eg the last layer of the conv2\_x block of ResNets \cite{He_2016_CVPR}). FPN gives an effective solution to explore the characteristic and advantage of the feature pyramid. Nevertheless, FPN holds an information flow from top to down, so the features at high levels are short of semantic available at low-level ones. The perspective of FPN inspires the follow-up researches on the exploration of building up better architectures to deal with multi-scale features. \cite{Liu_2018_CVPR, Ghiasi_2019_CVPR} introduce extra but limited information from other scales beyond the top-down path of FPN. To a certain degree, the above methods mitigate the scale variance problem existing in FPN, but ignore the channel and spatial semantic differences between the multi-scale features.

Inspired by the above researches, we propose the multi-scale convolution (MSConv) to effectively solve the mentioned problem. MSConv is an extension of traditional convolution to process multi-scale input. MSConv is computationally efficient and effective to improve the detection performance of the object detectors.

In this paper, we mainly make the following contributions.
\begin{itemize}
    \item[1.] We propose the multi-scale convolution (MSConv), to extend the traditional convolution to accept multi-scale input. MSConv can deal with the multi-scale features across scale, channel and spatial dimension at the same time.
    \item[2.] By replacing the convolutions with MSConvs in the detection head, mainstream single-stage object detectors can get more than 2.5\% improvement in AP. Our best model based on FreeAnchor \cite{zhang2019freeanchor} detector achieves 48.9\% AP on COCO \textit{test-dev} under single-scale testing, surpassing many state-of-the-art methods.
    \item[3.] The proposed MSConv is computationally efficient, \ie only a small increase of computation cost is required.
\end{itemize}

\section{Related Works}

\subsection{Object Detectors}
State-of-the-art object detection methods can usually be divided into two categories: single-stage detectors and two-stage ones.

\textbf{Two-stage Detectors.} To locate the objects with different scales, R-CNN systems \cite{2014Rich, girshickICCV15fastrcnn, renNIPS15fasterrcnn, cai18cascadercnn, chen2019hybrid} first generate region proposals at different scales, then extract the features within the region proposals for further classification or regression. Though the region proposals are at different scales, the extracted features are resized to the same spatial size (\eg $7 \times 7$) using ROI pooling \cite{girshickICCV15fastrcnn} or ROIAlign \cite{He_2017_ICCV} resizing operation. However, the scale variance problem still exists. The features extracted by the region proposals are from objects with different scales, containing the rich information on position and category of the objects. The larger objects with larger area have more spatial information than smaller ones. The resizing operation may bring inequality information loss for objects at different scales. As the features of any object are resized to the same spatial size, the information loss of larger objects is greater than that of the small ones. Therefore, there still exists scale variance problem in two-stage detectors.


\textbf{Single-stage Detectors.} Given multi-scale input features, single-stage detectors (\eg \cite{liu2016ssd}) usually directly generate predictions. Anchor-based RetinaNet \cite{Lin_2017_ICCV} has made great progress in solving scale variance problem. While dense anchors with different scales and aspect ratios are used by RetinaNet to cover different objects, FPN \cite{Lin_2017_CVPR} is used to enhance the features with stronger semantic from higher levels. RetinaNet has better performance than Faster R-CNN \cite{renNIPS15fasterrcnn} and comparable performance with most two-stage detectors. Anchor-free ones \cite{tian2019fcos, kong2019foveabox} adopt per-pixel prediction of location and classification. To avoid object ambiguity problem, each pixel is only related to single object. Objects will be assigned to the corresponding levels according to their scales. The larger objects are assigned to higher levels for their larger receptive fields, the smaller objects are assigned to lower levels for their finer spatial features. Therefore, each level is in charge of prediction of objects at similar scales. However, this fixed assignment strategy only associates any object with a certain scale, which is limited for object represent. It may be better to predict the object using features from more scales (\eg features that are semantically stronger or spatially finer), instead of the fixed ones.

\subsection{Methods Dealing with Scale Variance Problem}
Recently, many researchers are exploring methods to overcome the scale variance problem. We can simply divide them into two categories: methods fusing features from partial scales (partial connections for short) and methods fusing features from all scales (full connections for short).

\textbf{Partial Connections.} FPN \cite{Lin_2017_CVPR} first proposes lateral connection to merge features from adjacent scales in a top-down path. PANet \cite{Liu_2018_CVPR} brings an additional bottom-up path on the basis of FPN to supplement the missing finer spatial information from smaller scales. NAS-FPN \cite{Ghiasi_2019_CVPR} introduces neural architecture search (NAS) to discover a better scheme that merges cross-scale features from any scales instead of only adjacent ones. These methods enhance the original features with semantics from other scales, but can only obtain information from limited ones.

\textbf{Full Connections.} Beyond obtaining information from certain scales, many researches are exploring methods aggregating features from all scales. Kong \etal \cite{Kong_2018_ECCV} gather features from all levels to a medium level followed by concatenation along channel dimension. For each level, they use global attention and local configuration to enhance and refine the combined features. The refined features are finally resized to the corresponding level and then element-wisely summed up with the original input of this level. Libra R-CNN \cite{Pang_2019_CVPR} first gathers features from all levels to a medium level and does element-wise summation. After that, a non-local \cite{Wang_2018_CVPR} module is applied to enhance the merged features. The enhanced features are then scattered to each level and element-wisely summed up with the original input. The above methods obtain features from all scales by gathering them to a medium level and then merge the features. However, a feature representation at medium scale is improper to describe objects at other scales. Therefore, the generated features can fit well the scale of this level, but may not fit well for other scales.


Besides, most of the above methods adopt a simple way to merge the features (often element-wisely summation), which lacks nonlinearity and gives the features from different scales the same weight. Nevertheless, we should let the network learn proper weights to combine the features from different scales.

\section{Multi-Scale Convolution}

\subsection{Overview}


\noindent In this section, we give the overview of multi-scale convolution. Multi-scale convolution consists of two steps: feature gathering and feature blending. Let the input features from $L$ different levels be $X=\{X^1, \dots, X^L\}$.



In the feature gathering step, the multi-scale features will be gathered to each level. The output of feature gathering step $Q = \{Q^1, \dots, Q^L\}$ is obtained in a gather-scatter manner:


\begin{equation}
\label{eq:gather}
    \Phi = Gather(X, l_{gl}),\ l_{gl} \in \{1, \dots, L\}
\end{equation}

\begin{equation}
\label{eq:scatter}
    Q = Scatter(\Phi, \{l\}_{l=1}^L)
\end{equation}
where $\Phi$ denotes gathering result of all levels, $l_{gl}$ denotes the gathering level (set to 1 if not specified). \textit{Gather} and \textit{Scatter} denote the gathering and scattering process, respectively.

In the feature blending step, at each level, the gathered features are passed to two modules: scale alignment (SA) and context attention (CA), to further blend the multi-scale and original features. The output $O=\{O^1, \dots, O^L\}$ after CA can be computed as:
\begin{equation}
\label{eq:feature_blending}
    O = \{CA(W_m \ast SA(Q^l, X^l))\}_{l=1}^L
\end{equation}
where $W_m$ denotes the weight of the $1 \times 1$ convolution to make the channel number of the output of SA and $X^l$ equal. $\ast$ denotes the convolution operation.

The final output $Y=\{Y^1, \dots, Y^L\}$ of MSConv can be computed as:
\begin{equation}
\label{eq:output}
    Y = \{W_Y \ast (O^l \oplus X^l)\}_{l=1}^L
\end{equation}
where $\oplus$ denotes element-wise summation, $W_Y$ denotes the weight of the $3 \times 3$ convolution to generate the final output. 




\begin{figure}[h]
    \centering
    \includegraphics[width=0.46\textwidth]{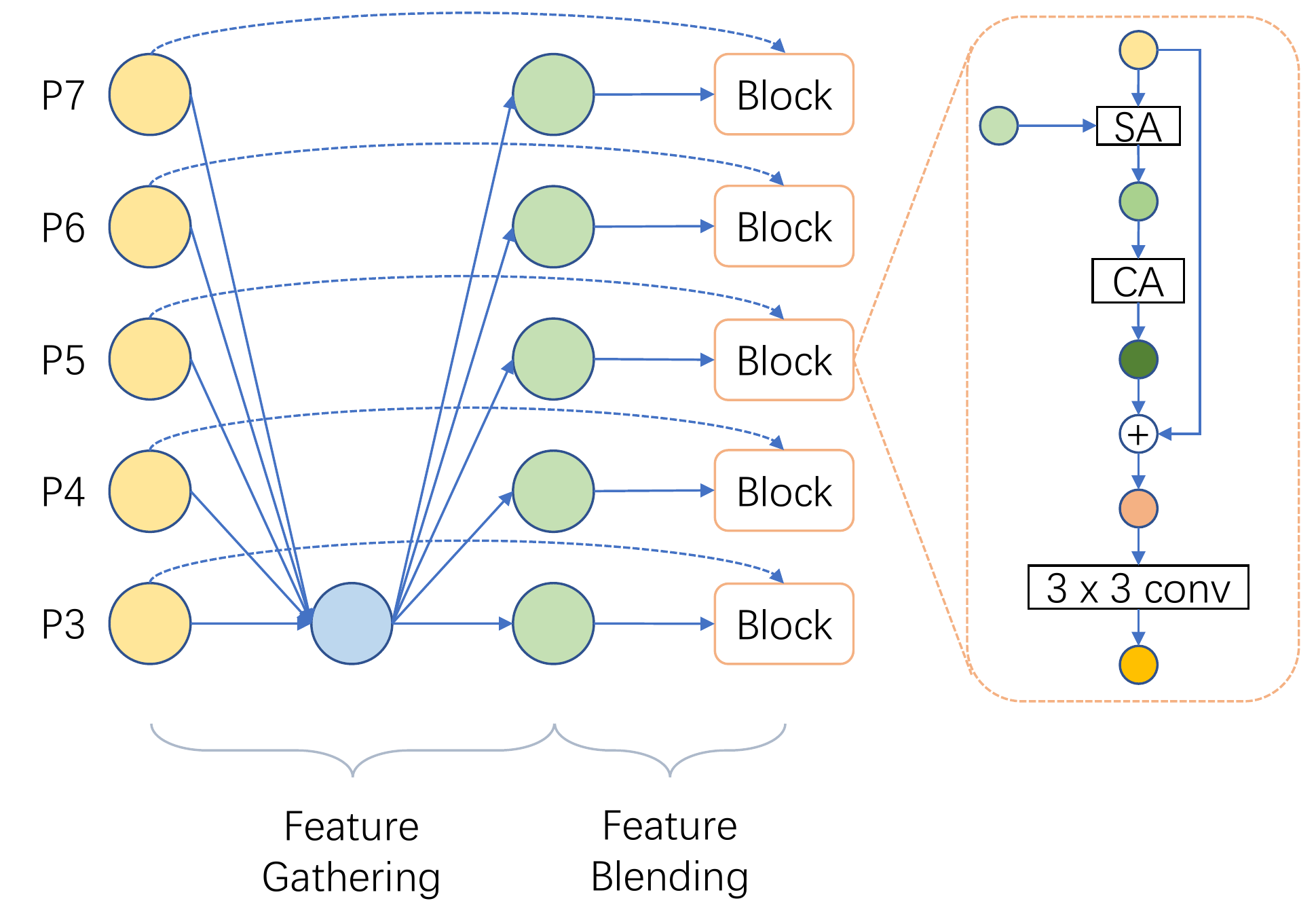}
    \caption{Architecture of the multi-scale convolution.}
    \label{fig:overview}
\end{figure} 


\subsection{Detailed Architecture}


Figure \ref{fig:overview} shows the architecture of our multi-scale convolution. In the feature gathering step, we first reduce the channels of the input features and resize them to the lowest level. Then we concatenate the gathered multi-scale features and scatter them to each level. In the feature blending step, the features at each level are then passed to a shared block for further processing. In the shared block, the multi-scale features with the original input features are passed to a scale alignment module to align the spatial scale of the multi-scale features. The scale-aligned features are then passed to a $1 \times 1$ convolution to merge the channels. The merged features are then rescaled by the context attention module with attention across scale, channel and spatial dimension. After that, a $3 \times 3$ convolution is applied in the element-wise summation of the merged features and the original input to generate the final output for each level.

\subsubsection{Feature Gathering}

\begin{figure}
    \centering
    \subfigure[Full connection.]{
        \includegraphics[width=0.194\textwidth]{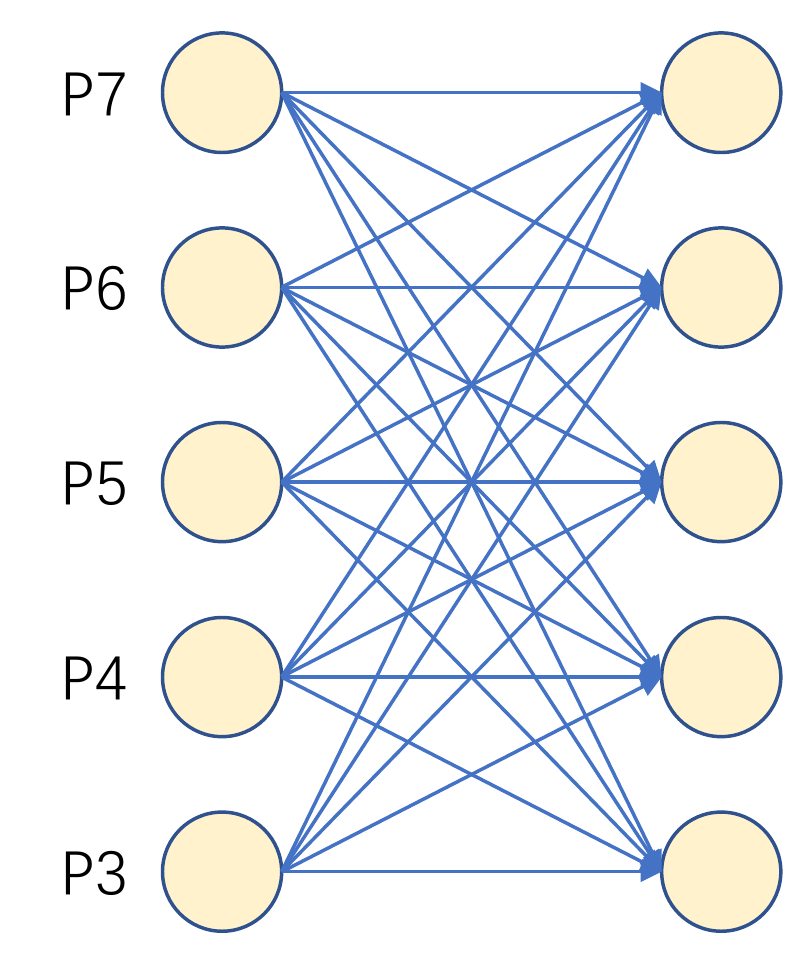}
    }
    \subfigure[Gather-scatter connection.]{
        \includegraphics[width=0.25\textwidth]{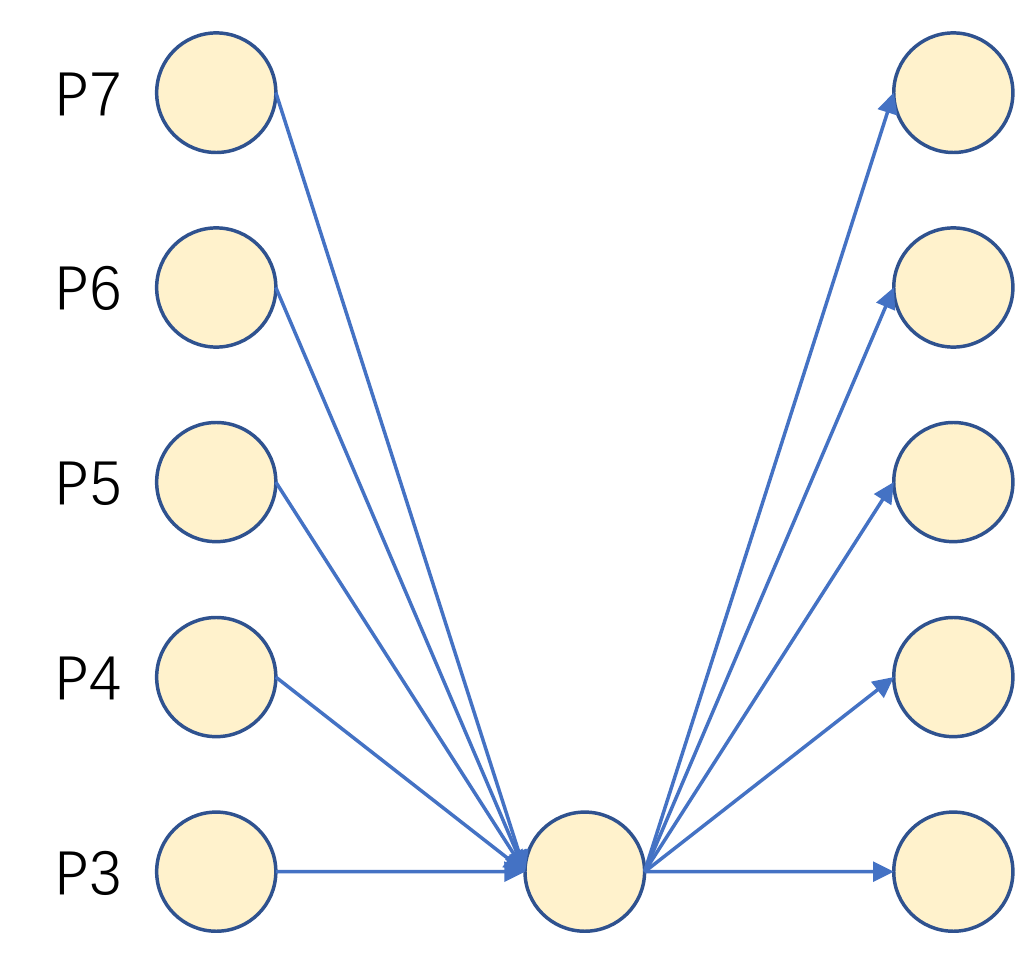}
    }
    \caption{Different connections of multi-scale features.}
    \label{fig:diff_connection}
\end{figure}

Before merging the multi-scale features from each scale, we should find a proper way of multi-scale feature representation for each level. The best solution is to gather features from all scales to each level, by full connection, shown in Figure \ref{fig:diff_connection} (a). In this way, the multi-scale features turn into the same spatial representation for each level. However, the full connection manner introduces too many additional operations including upsampling and downsampling, with a complexity of $\mathcal{O}(CL^2)$. As an alternative, we adopt a gather-scatter connection to approximate full connection. 

To reduce the computation cost, we first separately use a $1 \times 1$ convolution to reduce the number of channels of each input to $C_r$ 
($C_r \leq C$). The output of the $l$-th level is denoted as $D^l \in \mathbb{R}^{C_r \times H^l \times W^l}$, where $H^l$ and $W^l$ denotes the resolution of the input at the $l$-th level. $C_r$ is set to 64 in our experiment if not specified.
    
After that, to simultaneously process the multi-scale features, we gather all the features to the same level $l_{gl}$, and concatenate them along the channel dimension. $l_{gl}$ is set to 1 so that the gathered features can keep the largest resolution to avoid information loss. The gathering process to generate output $\Phi \in \mathbb{R}^{LC_r \times H^{l_{gl}} \times W^{l_{gl}}}$ is:

\begin{equation}
    E = \{Resize(D^l, (H^{l_{gl}}, W^{l_{gl}}))\}_{l=1}^L
\end{equation}
\begin{equation}
    \Phi = Concat(E)
\end{equation}
where $Resize$ denotes the resizing function, $Concat$ denotes concatenation along channels.

Then we generate the features for each level for further processing by scatter $\Phi$ to each level through resizing. The scatter process to generate $Q$ is:

\begin{equation}
    Q = \{Resize(\Phi, (H^l, W^l))\}_{l=1}^L
\end{equation}

In this way, for each level, the features from all scales obtain the same spatial scale of the current level, so that the detector can get a proper multi-scale feature representation, which can be processed across scale, channel and spatial dimension. The complexity of our connection manner is $\mathcal{O}(C_rL) < \mathcal{O}(CL^2)$, which is far less than that of the full connection manner.


\subsubsection{Feature Blending}
After feature gathering, we blend the features for each level. We first apply the scale alignment module to neutralize the spatial offset generated by the pooling operation during feature preparation. Then we merge the multi-scale features through a $1 \times 1$ convolution and use the context attention module to dynamically rescale the weight of each channel of the merged features. The network can thus dynamically select the useful features and suppress the useless features. At each level, the rescaled features are then element-wisely summed up with the original input and passed to a $3 \times 3$ convolution to generate the final output.

\begin{figure}
    \centering
    \includegraphics[width=0.42\textwidth]{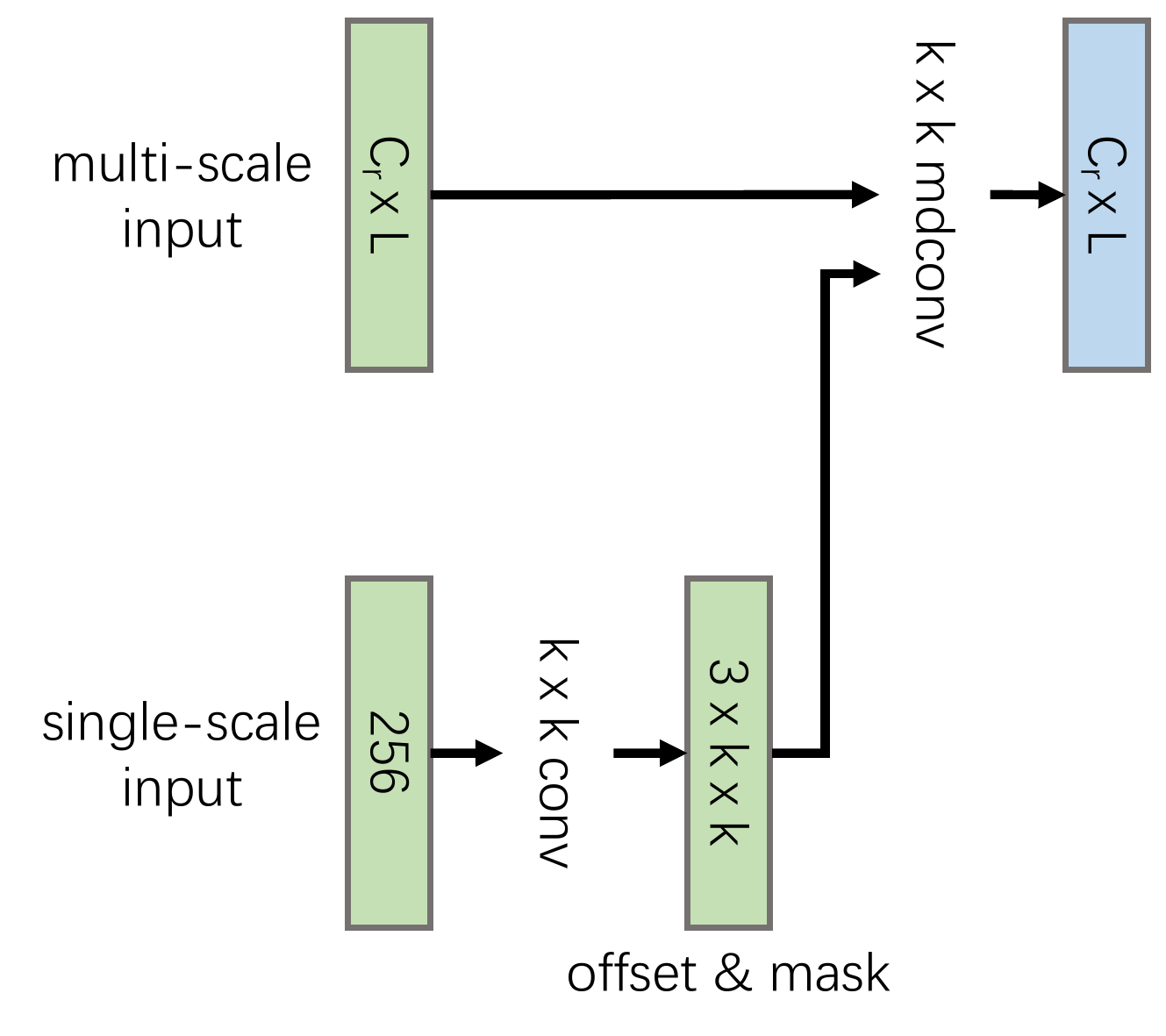}
    \caption{The scale alignment module. The rectangle represents a feature, in which the number of its channels is shown.``k" denotes the kernel size.}
    \label{fig:sa}
\end{figure}

\indent \textbf{Scale Alignment.} The pooling operation we use during feature preparation has translation invariance, and is not sensitive to position variation. Therefore, each pixel of the multi-scale features after feature gathering for each level has a spatial offset from the concatenated features $\Phi$. To deal with this problem, we propose the scale alignment (SA) module to neutralize the spatial offset. Figure \ref{fig:sa} shows the architecture of SA module. Firstly, we use a $k \times k$ convolution on the original single-scale input to generate the deformable offset and mask on the basis of the current scale. The multi-scale features with the offset and mask together are then passed to a $k \times k$ modulated deformable convolution \cite{Zhu_2019_CVPR_DCNv2} with groups=$L$, to generate the scale-aligned features $M^l$ at level $l$ ($k$ is set to 1 if not specified). Then a $1 \times 1$ convolution is applied in the multi-scale features to merge the channels. The merges features $M^l$ has the same channel number as $X^l$.

\begin{figure}
    \centering
    \includegraphics[width=0.46\textwidth]{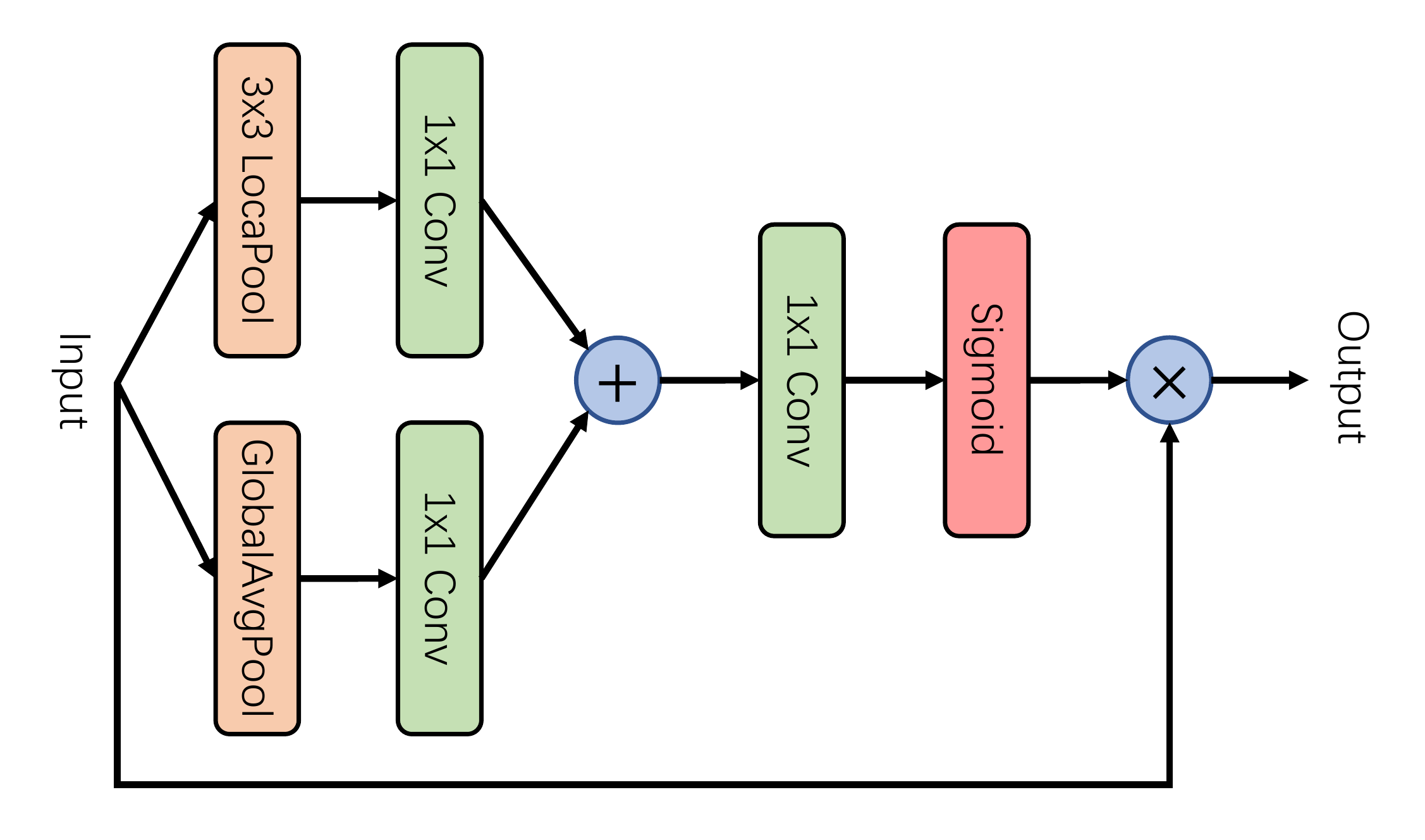}
    \caption{The context attention module.}
    \label{fig:ca}
\end{figure}

\textbf{Context Attention.} The merged features $M^l$ contains different context both in spatial and depth dimension. As the unnecessary context may bring influence to the final feature representation, we should keep the useful features and suppress the useless ones. In order to achieve this goal, we propose the context attention (CA) module to rescale the features by an attention across depth and spatial dimension. The architecture of CA module is shown in Figure \ref{fig:ca}. We first independently use a $3 \times 3$ local average pooling (LAP) and a global average pooling (GAP) to extract local features $L^l$ and global features $G^l$:

\begin{equation}
    L^l = LAP(M^l),\ G^l = GAP(M^l)
\end{equation}

Then a dependent $1 \times 1$ convolution is separately applied to $L^l$ and $G^l$ to extract features across channels:


\begin{equation}
    \mathcal{L}^l = W_{\mathcal{L}} \ast L^l,\ \mathcal{G}^l = W_{\mathcal{G}} \ast G^l
\end{equation}

\noindent where $\mathcal{L}^l$ and $\mathcal{G}^l$ are the features generated from $L^l$ and $G^l$, respectively. $W_{\mathcal{L}}$ and $W_{\mathcal{G}}$ are the weights of the convolution of $L^l$ and $G^l$ shared across all levels, respectively.

To generate the attention $S^l$ across depth (exactly across scale and channel) and spatial dimension, we apply a $1 \times 1$ convolution followed by a \textit{sigmoid} function on the element-wise summation of $\mathcal{L}^l$ and $\mathcal{G}^l$:

\begin{equation}
    S^l = \sigma(W_{\sigma} \ast (\mathcal{L}^l \oplus \mathcal{G}^l))
\end{equation}

\noindent where $W_\sigma$ denotes the weight of the convolution before sigmoid function, and $\sigma$ denotes the sigmoid function.

The output of CA module $O^l$ is an element-wise product of the merged features $M^l$ and the attention $S^l$:

\begin{equation}
    O^l = M^l \otimes S^l
\end{equation}

\begin{figure}
    \centering
    \subfigure[RetinaNet head with traditional convolutions.]{
        \includegraphics[width=0.46\textwidth]{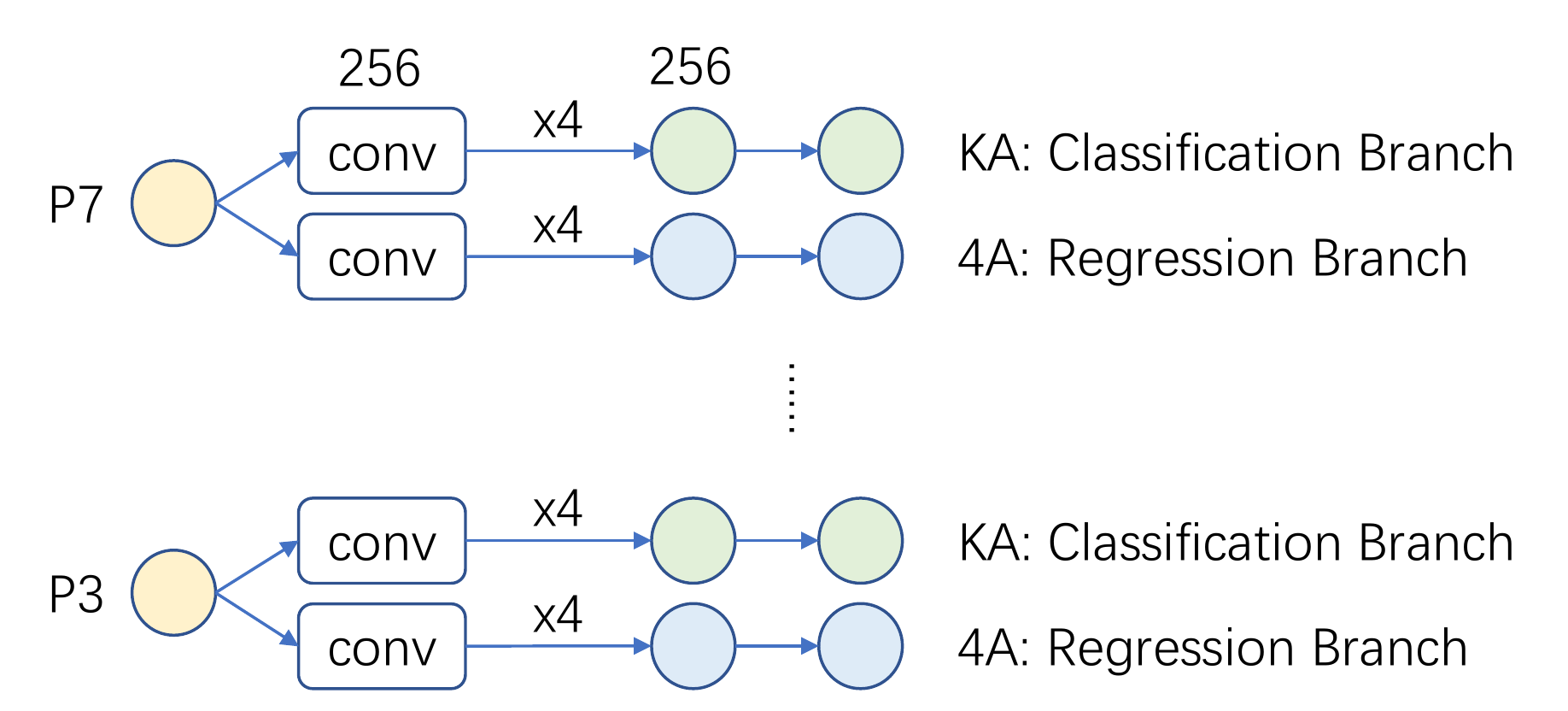}
    } \\ 
    \subfigure[RetinaNet head with multi-scale convolutions.]{
        \includegraphics[width=0.46\textwidth]{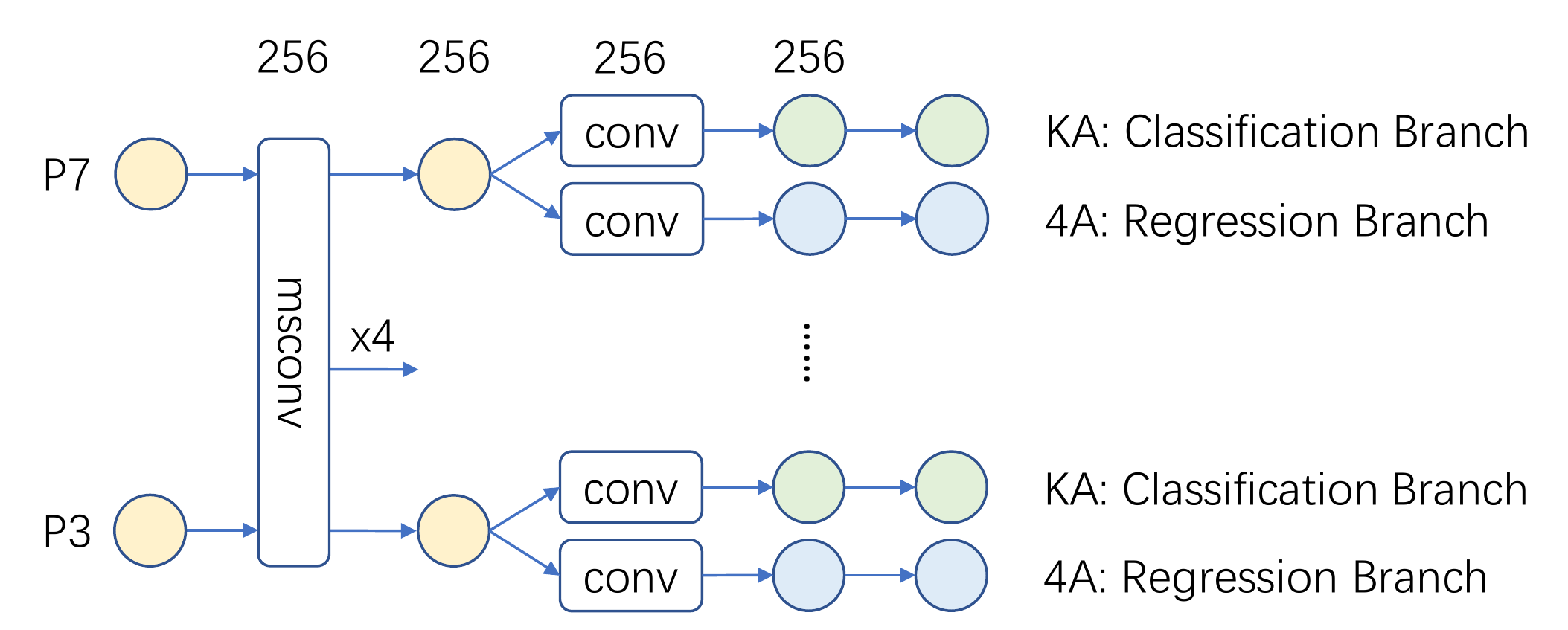}
    }
    \caption{RetinaNet head with different convolutions.}
    \label{fig:diff_conv_head}
\end{figure}

\begin{table*}
\centering
\caption{Ablation studies of component effectiveness on COCO \textit{val-2017}.``SA" and ``CA" denote scale alignment and context attention module, respectively. The number in [] is the relative improvement over RetinaNet baseline.}
\renewcommand\arraystretch{1.5}
\setlength{\tabcolsep}{2mm}{
\begin{tabular}{ccccccccccc}
\hline
 &  & $AP$ & $AP_{50}$ & $AP_{75}$ & $AP_S$ & $AP_M$ & $AP_L$ & Params (M) & FLOPs (G)\\ 
\hline
\multicolumn{2}{c}{RetinaNet baseline} & 33.2 & 52.5 & 35.4 & 15.8 & 37.1 & 46.5 & 37.74 & 95.56\\ 
\hline 
\hline 
\multicolumn{2}{c}{Ours}                    &  \\ 
SA & CA & $AP$ & $AP_{50}$ & $AP_{75}$ & $AP_S$ & $AP_M$ & $AP_L$ & Params (M) & FLOPs (G)\\ 
\hline 
           &            & 34.8[+1.6] & 55.1 & 37.2 & 17.0 & 38.8 & 48.8 & 37.60 & 93.20 \\ 
\checkmark &            & 35.6[+2.4] & 56.1 & 38.2 & 18.7 & 39.9 & 50.3 & 37.70 & 94.13 \\ 
           & \checkmark & 35.6[+2.4] & 56.2 & 38.1 & 18.3 & 39.9 & 49.2 & 38.39 & 97.69 \\ 
\checkmark & \checkmark & \textbf{35.9}[\textbf{+2.7}] & \textbf{56.4} & \textbf{38.9} & \textbf{18.4} & \textbf{40.0} & \textbf{50.6} & 38.49 & 98.61 \\ 
\hline \\
\end{tabular}
}
\label{table:ablation}
\end{table*}

\begin{figure*}
    \centering
    \subfigure[Original image.]{
        \includegraphics[width=0.25\textwidth]{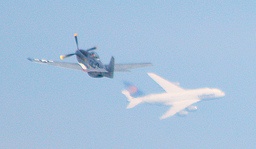}
    } \quad 
    \subfigure[Score map w/o SA.]{
        \includegraphics[width=0.25\textwidth]{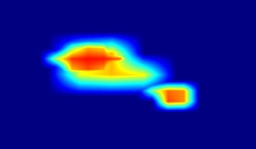}
    } \quad
    \subfigure[Score map w/ SA.]{
        \includegraphics[width=0.25\textwidth]{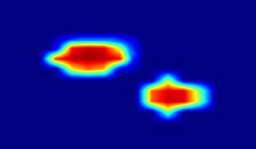}
    }
    \caption{Visualization of the confidence score maps without or with scale alignment module.}
    \label{fig:sa_comp}
\end{figure*}

\begin{table}[]
\centering
\caption{The effect of different gathering levels on COCO \textit{val-2017}.}
\renewcommand\arraystretch{1.5}
\renewcommand\tabcolsep{1.3mm}
\begin{tabular}{cccccccccc}
\hline
Gathering Level & $AP$ & $AP_{50}$ & $AP_{75}$ & $AP_{S}$ & $AP_{M}$ & $AP_{L}$\\ \hline
P7 & 34.6 & 55.2 & 37.0 & 17.6 & 38.6 & 48.1 \\ 
P5 & 35.2 & 55.7 & 37.7 & 17.4 & 39.1 & 49.2 \\ 
P3 & \textbf{35.9} & \textbf{56.4} & \textbf{38.9} & \textbf{18.4} & \textbf{40.0} & \textbf{50.6} \\ 
\hline \\
\end{tabular}
\label{table:gather_level}
\end{table}

\begin{table*}[]
\centering
\caption{Comparisons with other pyramid architectures based on RetinaNet detector on COCO \textit{val-2017}. The number in [] is the relative improvement over FPN \cite{Lin_2017_CVPR}.}
\renewcommand\arraystretch{1.5}
\renewcommand\tabcolsep{2.8mm}
\begin{tabular}{lccccccccc}
\hline
Method & $AP$ & $AP_{50}$ & $AP_{75}$ & $AP_{S}$ & $AP_{M}$ & $AP_{L}$ & Params (M) & FLOPs (G) \\ \hline
FPN \cite{Lin_2017_CVPR} & 33.2 & 52.5 & 35.4 & 15.8 & 37.1 & 46.5 & \textbf{37.74}(\textbf{1.00x}) & 95.56(1.00x) \\ 
PANet \cite{Liu_2018_CVPR} & 33.4[+0.2] & 52.5 & 35.4 & 16.0 & 37.7 & 46.9 & 40.10(1.06x) & 97.92(1.02x) \\ 
PConv \cite{wang2020SEPC} & 33.8[+0.6] & 53.8 & 36.1 & 16.7 & 38.1 & 46.9 & 41.28(1.09x) & 96.78(1.01x) \\
Libra \cite{Pang_2019_CVPR} & 33.9[+0.7] & 53.8 & 36.1 & 16.5 & 38.0 & 47.6 & 38.01(1.01x) & 95.75(1.00x) \\ 

NAS-FPN \cite{Ghiasi_2019_CVPR} & 35.1[+1.9] & 53.9 & 37.3 & 17.1 & 39.7 & 49.8 & 59.72(1.58x) & 138.60(1.45x) \\  
BiFPN \cite{Tan_2020_CVPR_EfficientDet} & 35.2[+2.0] & 54.4 & 37.7 & 17.5 & 39.5 & 49.3 & 55.60(1.47x) & 122.34(1.28x) \\ 
SEPC-Lite \cite{wang2020SEPC} & 35.3[+2.1] & 55.3 & 37.6 & 17.6 & 39.4 & 49.9 & 41.37(1.10x) & 96.99(1.01x) \\ 
\hline
MSConv & \textbf{35.9}[\textbf{+2.7}] & \textbf{56.4} & \textbf{38.9} & \textbf{18.4} & \textbf{40.0} & \textbf{50.6} & 38.49(1.02x) & 98.61(1.03x) \\ 
\hline \\
\end{tabular}
\label{table:feature-fusion}
\end{table*}

\subsubsection{Head Design}

In this section, we introduce how to integrate our MSConv into single-stage detectors. We take RetinaNet as an example to elaborate how to replace the traditional convolution used in the detection head of single-stage detectors. Figure \ref{fig:diff_conv_head} shows the difference between head design of traditional convolutions and that of our MSConvs.

In the original RetinaNet, the multi-scale inputs are separately processed by a shared head with two branches: classification and regression. The two branches have independent weights but share the same input. At each branch, the input features are passed through a fully convolution network (FCN) consisting of several (4 by default) stacked convolutions to extract features specially for classification or regression. Finally, a $3 \times 3$ convolution is applied to the extracted features to generate the final prediction.

It is easily to replace the traditional convolutions with MSConvs at each branch. However, each MSConv still brings additional computation. To make a compromise, the two branches share the same MSConv. To keep the difference between classification and regression, we introduce an extra $3 \times 3$ convolution for each branch after the shared MSConvs. The final prediction at each branch is generated by a $3 \times 3$ convolution.

\section{Experiments}

\subsection{Dataset and Evaluation Metrics}
We carry out our experiments on COCO \cite{10.1007/978-3-319-10602-1_48} dataset. We use the data in \textit{train-2017} split containing around 115k images to train our models, and evaluate the performance for ablation studies on \textit{val-2017} split with about 5k images. Main results are reported on the \textit{test-dev} split (20k images without available public annotations). We report all the results in the standard COCO-style average precision (AP).

\subsection{Experimental Settings}
For fair comparisons, we conduct our experiments on MMDetecion \cite{mmdetection} platform in PyTorch \cite{NEURIPS2019_9015} framework. If not specified, all the settings are the same with described in MMDetection \cite{mmdetection}. Modulated deformable convolution \cite{Zhu_2019_CVPR_DCNv2} is applied.

\textbf{Training Settings.} We adopt ResNet-50 \cite{He_2016_CVPR} as our default backbone network, and RetianNet \cite{Lin_2017_ICCV} as our default object detector. The backbone networks are initialized with the weight of the models pretrained on ImageNet \cite{5206848}. Our models are trained using stochastic gradient descent (SGD) optimizer for 12 epochs with an initial learning rate of 0.01, which is divided by 10 after 8 and 11 epochs. Due to memory limitation, the batchsize (16 by default) will be adjusted with a linearly scaled learning rate. Momentum and weight decay are set to 0.9 and $1e ^ {-4}$, respectively. The resolution of the input images is set to $640 \times 640$.

\textbf{Inference Settings.} For each input image, we execute the following steps to get the predictions. We collect the predictions with top 1000 confidences from each prediction layer and use a confidence threshold of 0.05 for each class to filter out the predictions with low confidences. For each class, we apply non maximum suppression (NMS) with threshold of 0.5 to filter the low-quality predictions. After NMS, we select the predictions with top 100 confidences as the final results.

\subsection{Ablation Studies}

\subsubsection{The effectiveness of each component}

\indent We analyze if each component of our model is effective for improvement of detection. The experimental results are listed in Table \ref{table:ablation}. The performance of RetinaNet is shown in the first group, while the performance of our methods are shown in the second group.

The third line of the second group in the table shows that our plain model without any extra module can simply achieve a better performance than RetinaNet baseline (+1.6\% AP), even with fewer parameters and less computational cost.

If we only introduce the scale alignment module, the detector can get 0.8\% AP improvement over our plain model, which is 2.4\% higher than the original RetinaNet. As shown in Figure \ref{fig:sa_comp}, combined with the SA module, the foreground regions of the objects are more accurate. Introducing the context attention module, the detector gets an improvement of 2.4\% AP over RetinaNet baseline, with a little increase of computation.

The last line of the table reveals that applying both SA and CA module can boost the performance of the detector to 35.9\% AP,\ which is 2.7\% higher than that of the baseline. 

\subsubsection{The effectiveness of different gathering levels}

\indent In this section, we analyze the effect of different gathering level of MSConv. As shown in Table \ref{table:gather_level}, as the gathering level goes down (from P7 to P3), the performances under all metrics goes higher. The results show that the level with the largest scale (P3) can keep the most information with the least loss, which justifies the effectiveness to gather multi-scale features to the level with the largest resolution.


\begin{table}[]
\centering
\caption{Applied in single-stage and two-stage detectors on COCO \textit{val-2017}. The number in [] is the relative improvement over the original detector.}
\renewcommand\arraystretch{1.5}
\renewcommand\tabcolsep{2.0mm}
\begin{tabular}{ccccc}
\hline
Detector & MSConv & $AP$ & $AP_{50}$ & $AP_{75}$ \\ \hline
\textit{single-stages} \\ 
RetinaNet &  & 33.2 & 52.5 & 35.4 \\ 
RetinaNet & \checkmark & \textbf{35.9}[\textbf{+2.7}] & 56.4 & 38.9 \\ 
\hline
FreeAnchor & & 36.4 & 54.6 & 39.1  \\ 
FreeAnchor & \checkmark & \textbf{39.0}[\textbf{+2.6}] & 58.1 & 42.1 \\ 
\hline
RepPoints &   & 35.8 & 56.2 & 37.9 \\ 
RepPoints & \checkmark & \textbf{37.5}[\textbf{+1.7}] & 57.6 & 40.4  \\ 
\hline
\hline
\textit{two-stages} \\ 
Faster R-CNN &  & 34.6 & 55.7 & 36.9 \\ 
Faster R-CNN & \checkmark & \textbf{37.2}[\textbf{+2.6}] & 57.8 & 39.9  \\ 
\hline
Mask R-CNN &  & 35.2 & 56.4 & 37.9 \\ 
Mask R-CNN & \checkmark & \textbf{38.2}[\textbf{+3.0}] & 57.6 & 43.0 \\ 
\hline
Cascade R-CNN &   & 38.1 & 55.9 & 41.1 \\ 
Cascade R-CNN & \checkmark & \textbf{39.7}[\textbf{+1.6}] & 57.6 & 43.0  \\ 
\hline \\
\end{tabular}
\label{table:application}
\end{table}




\begin{table*}[!h]
\centering
\caption{Comparisons with state-of-the-art methods on COCO \textit{test-dev} under single model and single-scale testing settings.}
\renewcommand\arraystretch{1.4}
\renewcommand\tabcolsep{5mm}
\scriptsize{
\begin{tabular}{llccccccc}
\hline
Method & Backbone & $AP$ & $AP_{50}$ & $AP_{75}$ & $AP_{S}$ & $AP_{M}$ & $AP_{L}$ \\ \hline
\textit{Two-stage methods} & & \\
Faster R-CNN w/ FPN \cite{Lin_2017_CVPR} & ResNet-101 & 36.2 & 59.1 & 39.0 & 18.2 & 39.0 & 48.2 \\ 
Mask R-CNN \cite{He_2017_ICCV} & ResNet-101 & 38.2 & 60.3 & 41.7 & 20.1 & 41.1 & 50.2 \\ 
Mask R-CNN \cite{He_2017_ICCV} & ResNeXt-101 & 39.8 & 62.3 & 43.4 & 22.1 & 43.2 & 51.2 \\ 
LH R-CNN \cite{DBLP:journals/corr/abs-1711-07264} & ResNet-101 & 41.5 & - & - & 25.2 & 45.3 & 53.1 \\ 
Cascade R-CNN \cite{cai18cascadercnn} & ResNet-101 & 42.8 & 62.1 & 46.3 & 23.7 & 45.5 & 55.2 \\ 
TridentNet \cite{li2019scale} & ResNet-101 & 42.7 & 63.6 & 46.5 & 23.9 & 46.6 & 56.6 \\ 
TridentNet \cite{li2019scale} & ResNet-101-DCN & 46.8 & 67.6 & 51.5 & 28.0 & 51.2 & 60.5 \\ 
TSD \cite{song2020revisiting} & ResNet-101 & 43.2 & 64.0 & 46.9 & 24.0 & 46.3 & 55.8 \\ 
\hline
\textit{One-stage methods} & & \\
RetinaNet \cite{Lin_2017_ICCV} & ResNet-101 & 39.1 & 59.1 & 42.3 & 21.8 & 42.7 & 50.2 \\
RetinaNet \cite{Lin_2017_ICCV} & ResNeXt-101 & 40.8 & 61.1 & 44.1 & 24.1 & 44.2 & 51.2 \\ 
FreeAnchor \cite{zhang2019freeanchor} & ResNet-101 & 43.1 & 62.2 & 46.4 & 24.5 & 46.1 & 54.8 \\
FreeAnchor \cite{zhang2019freeanchor} & ResNeXt-101 & 44.9 & 64.3 & 48.5 & 26.8 & 48.3 & 55.9 \\ 
FCOS \cite{tian2019fcos} & ResNet-101 & 41.5 & 60.7 & 45.0 & 24.4 & 44.8 & 51.6 \\
FCOS \cite{tian2019fcos} & ResNeXt-101 & 44.7 & 64.1 & 48.4 & 27.6 & 47.5 & 55.6 \\ 
ATSS \cite{zhang2020bridging} & ResNet-101 & 43.6 & 62.1 & 47.4 & 26.1 & 47.0 & 53.6 \\ 
ATSS \cite{zhang2020bridging} & ResNet-101-DCN & 46.3 & 64.7 & 50.4 & 27.7 & 49.8 & 58.4 \\ 
ATSS \cite{zhang2020bridging} & ResNeXt-101-DCN & 47.7 & 66.6 & 52.1 & 29.3 & 50.8 & 59.7 \\ 
SAPD \cite{zhu2019soft} & ResNet-101 & 43.5 & 63.6 & 46.5 & 24.9 & 46.8 & 54.6 \\ 
SAPD \cite{zhu2019soft} & ResNet-101-DCN & 46.0 & 65.9 & 49.6 & 26.3 & 49.2 & 59.6 \\ 
SAPD \cite{zhu2019soft} & ResNeXt-101-DCN & 46.6 & 66.6 & 50.0 & 27.3 & 49.7 & 60.7 \\ 
RepPoints v2 \cite{chen2020reppoints} & ResNeXt-101 & 47.8 & 67.3 & 51.7 & 29.3 & 50.7 & 59.5 \\ 
RepPoints v2 \cite{chen2020reppoints} & ResNet-101-DCN & 48.1 & 67.5 & 51.8 & 28.7 & 50.9 & \textbf{60.8} \\ 
GFL \cite{li2020generalized} & ResNet-101 & 45.0 & 63.7 & 48.9 & 27.2 & 48.8 & 54.5\\ 
GFL \cite{li2020generalized} & ResNet-101-DCN & 47.3 & 66.3 & 51.4 & 28.0 & 51.1 & 59.2 \\ 
GFL \cite{li2020generalized} & ResNeXt-101-DCN & 48.2 & 67.4 & 52.6 & 29.2 & 51.7 & 60.2 \\ 
\hline 
FreeAnchor w/ MSConv & ResNet-101-DCN & 47.7 & 67.5 & 52.2 & 29.6 & 51.2 & 58.5 \\ 
FreeAnchor w/ MSConv & ResNeXt-101-DCN & \textbf{48.9} & \textbf{68.8} & \textbf{53.4} & \textbf{31.8} & \textbf{51.8} & 60.2 \\ 
\hline \\ 
\end{tabular}}
\label{table:comp_sota}
\end{table*}

\subsection{Comparisons with other Pyramid Architectures}
In order to justify that our model is more effective and efficient , we compare the performance of our method with state-of-the-art pyramid architectures in Table \ref{table:feature-fusion} based on RetinaNet detector. Though NAS-FPN \cite{Ghiasi_2019_CVPR} and BiFPN \cite{Tan_2020_CVPR_EfficientDet} achieve huge improvement over FPN \cite{Lin_2017_CVPR}, they require much more computational cost. 
The last line of the table shows that our model achieves the best performance among the state-of-the-art methods, with only a small increase of parameters and FLOPs.

\subsection{Application in Single-stage and Two-stage Detectors}

In this section, we conduct experiments to evaluate the effectiveness of our model on single-stage detectors, including anchor-based ones or anchor-free ones, such as FreeAnchor \cite{zhang2019freeanchor} and RepPoints \cite{yang2019reppoints}, and two-stage detectors. 

As shown in the first group of Table \ref{table:application}, when combined with our method, the single-stage detectors can get a significant improvement in AP. MSConv can provide an increase of larger than 2.5\% AP for the single-stage detectors. The comparison of MSConv in various single-stage detectors is also shown in Figure \ref{fig:performance}.

In additional to single-stage detectors, MSConv is also effective for two-stage detectors. 
The second group in Table \ref{table:application} lists the experimental results of MSConv applied to two-stage detectors. When combined with MSConvs, two-stage detectors get more than 2.5\% improvement in AP. MSConv provides the most increase in AP of 3.0\% for Mask R-CNN.

\subsection{Comparison with State-of-the-art Methods}


In this section, we compare the performance of our method on COCO \textit{test-dev} split with state-of-the-art methods under single model single-scale testing settings, which are shown in Table \ref{table:comp_sota}. We use FreeAnchor as our detector. For training, we adopt 2$\times$ learning schedule with scale-jitter. With ResNet-101-DCN backbone, the AP (47.7\%) of our method surpasses most of the state-of-the-art methods using the same backbone. 
The backbone of ResNeXt-101-DCN further improves our AP to 48.9\%, which surpasses the AP of all other competitors.


\section{Conclusions}
In this paper, we propose the multi-scale convolution (MSConv), an extension of the traditional convolution, to accept and deal with multi-scale input. The proposed MSConv can meanwhile process the multi-scale input across channel, spatial and scale dimension. MSConv can dramatically boost the detection performance of single-stage object detectors with only a small increase of computation. The results also suggest that MSConv is flexible and able to bring considerable improvement as well to two-stage object detectors. 

{\small
\bibliographystyle{IEEEtran}
\bibliography{References}
}

\end{document}